\documentclass[letterpaper, 10 pt, journal, twoside]{ieeetran}

\usepackage[ruled]{algorithm2e}
\usepackage{algpseudocode}
\usepackage{graphics}
\usepackage{amsmath}
\DeclareMathOperator*{\argmax}{argmax}

\usepackage{amssymb}
\usepackage{epsfig} 
\usepackage{booktabs}
\usepackage{float}
\usepackage{url}
\usepackage{gensymb}
\usepackage{tikz}
\usepackage{color}
\definecolor{darkgreen}{RGB}{0,127,0}
\definecolor{darkblue}{RGB}{0,0,175}

\newcommand{\remindtext}[2]{{{#2}}}
\newcommand{\addedtext}[2]{{{#2}}}
\newcommand{\modifiedtext}[2]{{{#2}}}
\newcommand{\deletedtext}[2]{{}}

\newcount\Comments 
\usepackage{array,multirow,graphicx}
\usepackage{color}
\usepackage{wrapfig}
\definecolor{darkgreen}{rgb}{0,0.5,0}
\definecolor{purple}{rgb}{1,0,1}
\newcommand{\kibitz}[2]{\ifnum\Comments=1\textcolor{#1}{#2}\fi}

\begin{document}

\title{TANDEM: Learning Joint Exploration and Decision Making with Tactile Sensors}

\author{Jingxi Xu$^{1}$, Shuran Song$^{1}$ and Matei Ciocarlie$^{2}$%
\thanks{Manuscript received: February 24, 2022; Revised: June 18, 2022; Accepted: July 6, 2022.}
\thanks{This paper was recommended for publication by Ashis Banerjee upon evaluation of the Associate Editor and Reviewers' comments.}
\thanks{This work was supported in part by NSF grants CMMI-2037101, ECCS-2143601 and ONR grants N00014-19-1-2062, N00014-21-1-4010.} 
\thanks{$^{1}$Department of Computer Science, Columbia University, New York, NY 10027, USA.
        {\tt\footnotesize \{jxu, shurans\}@cs.columbia.edu}}%
\thanks{$^{2}$Department of Mechanical Engineering, Columbia University, New York, NY 10027, USA.
        {\tt\footnotesize matei.ciocarlie@columbia.edu}}%
\thanks{Digital Object Identifier (DOI): see top of this page.}
}

\markboth{IEEE Robotics and Automation Letters. Preprint Version. Accepted July, 2022}
{Xu \MakeLowercase{\textit{et al.}}: TANDEM: Learning Joint Exploration and Decision Making with Tactile Sensors} 


\maketitle

\begin{abstract}
Inspired by the human ability to perform complex manipulation in the complete absence of vision (like retrieving an object from a pocket), the robotic manipulation field is motivated to develop new methods for tactile-based object interaction. However, tactile sensing presents the challenge of being an active sensing modality: a touch sensor provides sparse, local data, and must be used in conjunction with effective exploration strategies in order to collect information. In this work, we focus on the process of guiding tactile exploration, and its interplay with task-related decision making. We propose TANDEM (TActile exploration aNd DEcision Making), an architecture to learn efficient exploration strategies in conjunction with decision making. Our approach is based on separate but co-trained modules for exploration and discrimination. We demonstrate this method on a tactile object recognition task, where a robot equipped with a touch sensor must explore and identify an object from a known set based on \modifiedtext{1-1}{binary contact signals} alone. TANDEM achieves higher accuracy with fewer actions than alternative methods and is also shown to be more robust to sensor noise. 
\end{abstract}


\section{Introduction}

\IEEEPARstart{T}{actile} sensing plays an important role for robots aiming to perform complicated manipulation tasks when vision is unavailable due to factors like occlusion, lighting, restricted workspace, etc. 
The ability of touch to provide useful information in the absence of vision is immediately clear in the case of human manipulation: we are able to search and manipulate efficiently inside of a bag or pocket without visual data. In particular, we have little problem in distinguishing between similar objects from tactile cues only.

However, a number of challenges remain before tactile sensing can be used with similar effectiveness by robotic manipulators. Fundamentally, touch is an active sensing modality, and individual tactile signals are very local and sparse. Guidance becomes critical: tactile sensors need to be physically moved by a robotic manipulator to obtain new signals, introducing additional costs for every sensor measurement. Without smart guidance, we can only blindly scan/grope on a surface~\cite{okamura2001feature, allen1985object} or continuously make repetitive and high amounts of contacts at tightly controlled positions~\cite{meier2011probabilistic, allen1988integrating, bierbaum2009grasp, gaston1984tactile, skiena1989problems}. These strategies are extremely inefficient and often incur prohibitively high costs and burdens. Furthermore, it is also important to have an intelligent way to rearrange or encode such local and sparse signals into a global representation.



\begin{figure}[t]
    \centering
    \includegraphics[width=\linewidth]{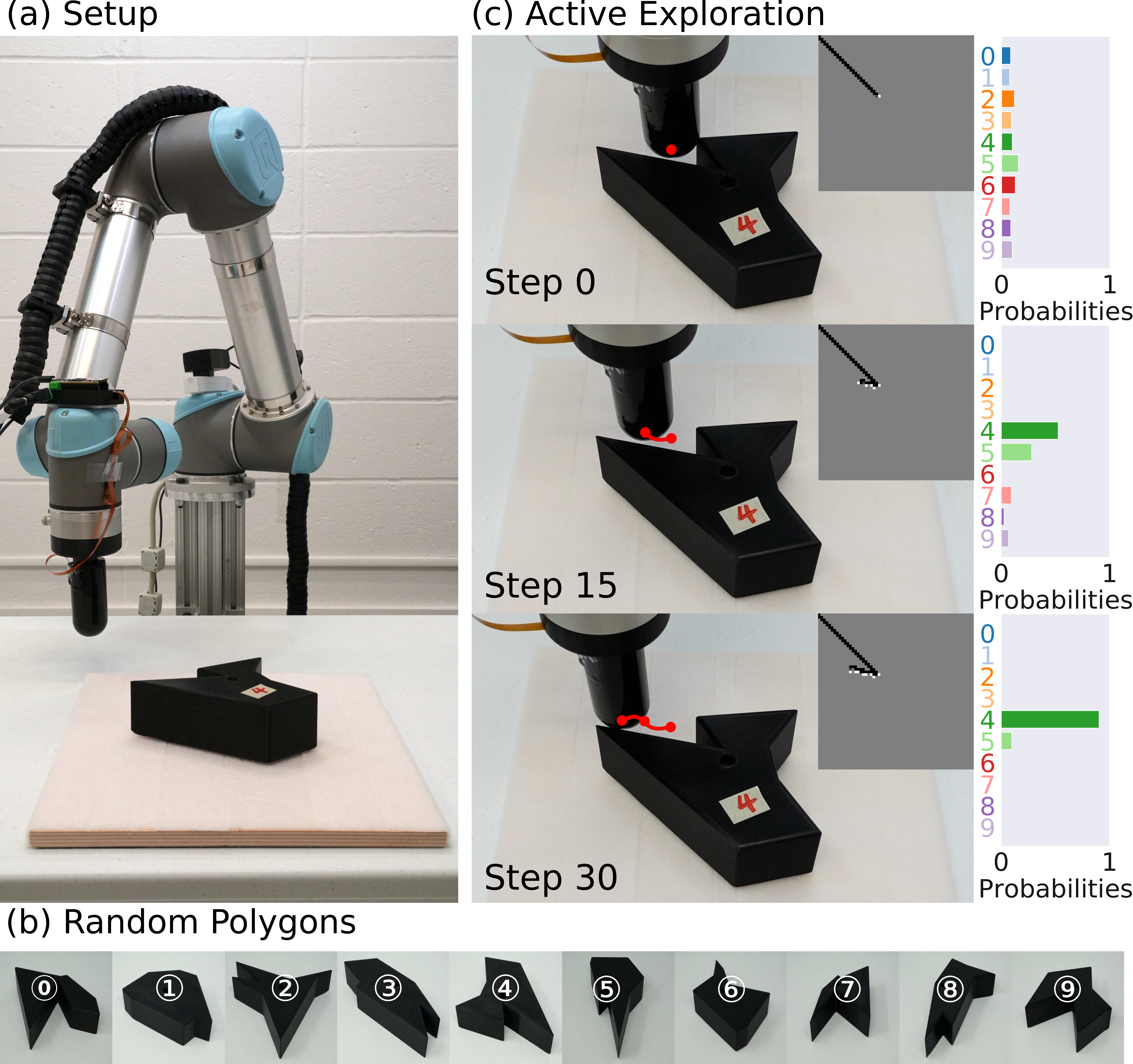}
    \caption{Object recognition based on tactile feedback alone. (a) Real robot setup. Our tactile finger is mounted on a robot arm, and the target object (unknown identity and orientation) is placed roughly around the workspace center. (b) Known object set of 10 randomly-generated polygons. (c) Active exploration. Using our framework, our robot collects data and quickly converges on the correct object identity (object 4 from the set).}
    \label{fig:teaser}
    \vspace{-0.15in}
\end{figure}

In this work, we focus on the process of guiding tactile exploration, and its interplay with task-related decision making. Our goal is to provide a method that can train effective guidance (exploration) strategies. The task we chose to highlight this interplay and to develop our method is tactile object recognition, in which one object must be identified out of a set of known models based only on touch feedback (Fig.~\ref{fig:teaser}). The goal of our method is to correctly recognize the object with as few actions as possible. 

In order to learn efficient guidance for such tasks, we propose an architecture combining an exploration strategy (i.e. \textit{explorer}) and a discrimination strategy (i.e. \textit{discriminator}). The \textit{explorer} guides the tactile exploration process by providing actions to take; the \textit{discriminator} attempts to identify the target object and determines when to terminate the exploration after enough information has been collected. To convert local and sparse tactile signals into a global representation, we also use an encoding strategy (i.e. \textit{encoder}). In our current version, the \textit{encoder} simply rearranges sparse tactile signals into an occupancy grid, but more complex implementations could be used for future tasks. 

In our proposed architecture, both the explorer and the discriminator are learned using data-driven methods; in particular, the explorer is trained via reinforcement learning (RL) and the discriminator is trained via supervised learning. \addedtext{1-1}{In our current implementation, both of these components are trained in simulation. The use of binary touch data, which is easier to simulate accurately compared to other tactile features, facilitates zero-shot sim-to-real transfer, which we demonstrate in the real robot experiments.}

Critically, even though our architecture separates the exploration and decision making, we interleave their training process: we propose a co-training framework that allows batch and repeated training of the discriminator on a set of samples collected by the explorer. We call our method \textbf{TANDEM}, for \textbf{TA}ctile exploration a\textbf{N}d \textbf{DE}cision \textbf{M}aking. In summary, the main contributions of this paper include:
\begin{itemize}
    \item We propose a new architecture to learn an efficient and active tactile exploration policy, comprising distinct modules for exploration, discrimination, and world encoding. We also propose a novel framework to co-train the exploration policy along with the task-related decision-making module, and show that they co-evolve and converge at the end of the training process.
    \item We demonstrate our method on a tactile object recognition task. In this context, we compare our approach against multiple baselines, including all-in-one learning-based approaches that do not distinguish between our proposed components, and other methods traditionally used for exploration (such as random-walk, info-gain, etc.) or tactile recognition (such as ICP). Our experiments, performed in simulation and validated on real robots, show that our proposed method outperforms these alternatives, achieving a higher success rate in identifying the correct object while also using fewer actions, and is robust to sensor noise.
\end{itemize}

\section{Related Work}

\subsection{Tactile Object Recognition}

Object recognition is a key problem in robotics and is a fundamental step to gaining information about the environment. Conventionally, visual perception has been the primary sensing modality for object recognition. 
However, due to the limitations of vision such as illumination and occlusion and with the development in tactile perception technology such as DISCO~\cite{piacenza2020sensorized},
object recognition with only tactile information is receiving increasingly wider attention in robotics research.

Tactile object recognition can be roughly divided into three major categories depending on the characteristics of the object~\cite{liu2017recent}: (1) rigid object recognition (the problem in this paper), (2) material recognition, and (3) deformable object recognition. 
However, many existing works are either using predefined action sequences or a heuristic-based exploration policy such as contour following, while we focus on developing a learning-based active exploration policy.


\subsection{Tactile Exploration Policy}
The Exploration Policy (EP) is the sequence of exploratory actions the agent executes to gather tactile information. Since tactile information can only be obtained by interacting with the target object, the EP plays a critical role. We divide tactile sensing EPs into three major categories.
\subsubsection{Passive mode} The robotic manipulator is fixed, and the human operator hands over the object to the manipulator, often times in random orientations and/or translations to collect tactile data~\cite{schmitz2014tactile, strub2014using, pastor2019using}.


\subsubsection{Semi-active mode} The manipulator interacts with the object according to a prescribed trajectory and does not need to react based on sensor data, maybe except being compliant to avoid damage. 
Examples include poking the object from uniformly sampled directions or grasping it multiple times with a predefined set of grasps~\cite{allen1988haptic, watkins2019multi, meier2011probabilistic}.

\begin{figure*}[t!]
\centering
\includegraphics[width=\textwidth]{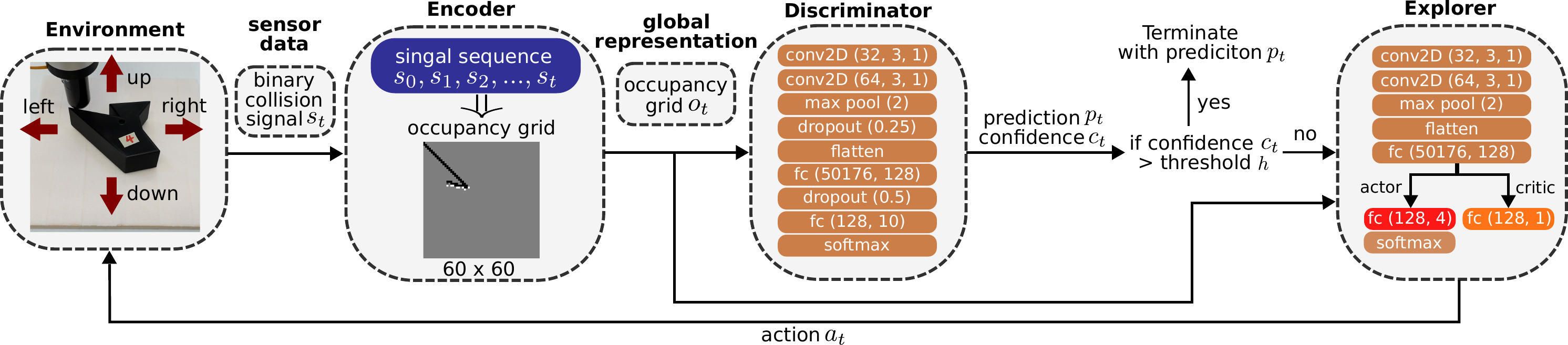}
\caption{An overview of the proposed architecture, and its application to tactile object recognition. The tactile finger interacts with the target object and generates local and sparse sensor data (in this task, binary collision signals). The encoder keeps a history buffer of such sequential signals and converts them into a global representation. Our encoder in this task rearranges them into an occupancy grid image. The discriminator takes in the global representation and attempts to identify the object along with a confidence estimate. If the confidence is higher than a predefined threshold, the exploration is terminated and the final prediction is produced. Otherwise, the explorer reads the representation and generates the next move. The neural networks used by the discriminator and explorer are shown inside their respective block. The parameters of the \texttt{conv2D} layer are the number of filters, kernel size, and stride. The parameters of the \texttt{max pool} layer is stride. The parameters of the \texttt{fc} layer are input dimension and output dimension. The parameter of \texttt{dropout} layer is the probability of an element being zeroed out.}
\label{fig:overview}
\vspace{-0.15in}
\end{figure*}

\subsubsection{Active mode} The manipulator finds the object and explores it reactively in a closed-loop fashion. The exploratory action is a function of current and/or past sensor data. EPs can be heuristic- or learning-based.

\modifiedtext{1-3}{Some of the most popular heuristic-based exploration policies range from contour following~\cite{martinez2013active, yu2015shape, suresh2020tactile, pezzementi2011tactile} to information gain (uncertainty reduction)~\cite{hebert2013next, xu2013tactile, schneider2009object, 
driess2017active}. 
Other heuristics to decide the regions of interest to explore include attention cubes~\cite{rajeswar2021touch}, Monte Carlo tree search~\cite{zhang2017active} and dynamic potential fields~\cite{bierbaum2009grasp}. However, while heuristic-based EPs require no training and can reduce the number of actions effectively, they are also sensitive to sensor noise and the performance of a particular heuristic can be task-dependent. In contrast, our learning-based EP is trained with sensor noise, and thus outperforms heuristic-based baselines when such noise is present in the evaluation.} 

\addedtext{1-3}{Similar to ours, other works combine exploration and decision making, whereby a classifier is pre-trained from  pre-collected data and used to estimate action quality with Bayesian methods to reduce uncertainty~\cite{fishel2012bayesian, lepora2013active, martinez2017active, kaboli2017tactile, kaboli2019tactile}. Most of these make effective use of high-dimensional or multimodal tactile data. Our use of relatively simple contact signals allows training an exploration policy through trial and error in simulation, with zero-shot transfer to real robots, eliminating the need for training on physical objects. Nevertheless, we achieve high recognition accuracy with relatively few actions, which we attribute in part to the fact that, unlike in previous methods, our discriminator is constantly updated as the exploration policy improves.}





\section{Architecture}



Our work aims to develop a framework that combines effective exploration and decision-making when using an active and local sensing modality, such as touch. Our key insight is that exploration and decision-making are distinct, yet deeply intertwined components of such a framework. An ideal exploration strategy will strive to reveal information that the decision-making component can make the best use of. Similarly, a decision-making component will adapt to the constraints of a real-world robot collecting touch data, which can only be obtained sequentially and incrementally. 

The concrete task we develop and test our method on is touch-only object recognition using a robot arm equipped with a tactile finger. We assume a set of known two-dimensional object shapes (randomly-generated polygons). One object is placed in the robot's workspace, in an unknown orientation. The robot must determine the object's identity using only tactile data, and with as little movement as possible. Performance is measured by both identification accuracy and the number of robot movements. 

Our proposed architecture is illustrated in Fig.~\ref{fig:overview}. The key components are the following: (1) The \textbf{explorer}, which generates an action for the robot to take in order to collect more data. In our implementation, the explorer consists of a policy trained via deep RL. (2) The \textbf{discriminator}, which predicts the identity of the object, along with a confidence value. This is a supervised learning problem, implemented in this case as a Convolutional Neural Network (CNN). Finally, in addition to the explorer and discriminator, we distinguish one additional component, namely (3) the \textbf{encoder} which converts the sequence of local and sparse tactile signals into a global representation. \remindtext{1-1a}{For our object recognition problem, the encoder simply aggregates binary touch signals into an occupancy grid.}

An equally important aspect of the proposed architecture is the training process. While we formulate distinct explorer and discriminator modules, trained via different formalisms (RL vs. supervised learning), we choose to interweave their training processes. This allows us to train the discriminator with data batches gathered by the explorer, which significantly improves data efficiency compared to an all-in-one approach that combines exploration and decision-making into a single component. In the co-training process, the explorer learns to increase the discriminator's confidence as fast as possible, and the discriminator learns to predict object identity based on the type of data generated by the explorer. 

\subsection{Encoder}

The job of the encoder is to maintain a history buffer of the sequence of contact data, convert that history into a global representation, and provide this representation as input to both the explorer and the discriminator. In our current implementation, we use binary signals indicating touch / no-touch. The encoder simply integrates these into an occupancy grid representation of the world, as shown in Fig.~\ref{fig:overview}. 

All pixels of the occupancy grid are initially grey (unexplored). After each action, if contact is detected, the corresponding pixel is colored white; otherwise, it is colored black. We also use a special value (light grey) to mark the current position of the finger on the grid. Knowing the current location of the finger is useful for the explorer to compute the next action; however, this special color is eliminated when the grid is provided as input to the discriminator because such information is not necessary for predicting the object identity.

For the task addressed here, we believe an occupancy grid works well due to its simple nature, ability to represent geometrical information, and small size in memory. However, when aggregating more complex information (e.g. from tactile sensors providing more than binary touch signals) or for more complex tasks, we expect that different encoding methods will be needed, even while the role in the architecture will be the same. We hope to explore more complex, learning-based encoders for our architecture in future studies.

\subsection{Discriminator}
The discriminator is the component of our pipeline in charge of interpreting sensor data for task-related purposes. Thus, for our problem, its job is to provide a prediction regarding the object identity, along with an associated confidence value. Making a confident prediction also implicitly terminates the exploration. 

In our implementation, underlying the discriminator is a CNN, as shown in Fig.~\ref{fig:overview}, taking as input the occupancy grid produced by the encoder. The network consists of two convolutional layers followed by a max-pool layer. After the dropout layer, the input is then flattened to go through another two fully-connected layers. A softmax function is applied to the raw 10-dimensional output from the fully-connected layer to generate a probability distribution. The object with the highest probability is chosen as the predicted identity and its corresponding probability is the confidence estimate. If the prediction confidence is greater than a preset threshold, the exploration is terminated and a final prediction is made. Otherwise, the occupancy grid is passed to the explorer to generate the next move.


\addedtext{1-6a}{As part of the co-training process, the discriminator is trained on partially complete occupancy grids, which can be ambiguous over objects, especially when very few pixels have been explored. This ambiguity is in fact the supervision needed to learn a confidence estimate. For instance, if the discriminator data buffer contains multiple duplicates of a highly incomplete grid, each with a different object label, then, in order to minimize the loss, the discriminator network will assign equal probabilities to all candidate objects, thus decreasing the confidence in each individual prediction.}

\subsection{Explorer}
The job of the explorer is to generate the next action for the robot, actively collecting additional information. For our task, this means selecting the next move (up, down, left, or right). Tactile data is collected automatically during the move and passed to the encoder as described above.

We implement the explorer as a Proximal Policy Optimization (PPO)~\cite{schulman2017proximal} agent taking the occupancy grid provided by the encoder as input. It has a similar architecture as the discriminator but the last fully-connected layer is replaced by a separate fully-connected layer for both the actor and critic, as shown in Fig.~\ref{fig:overview}. Even though the discriminator and explorer share part of the same architecture, we found through experiments that keeping the weights separate has a much better performance. This is likely because the discriminator and explorer focus on different aspects of the grid and should learn separate intermediate embeddings. As mentioned earlier, the grid input to the explorer has an extra bit of information providing the current location of the agent.

The reward structure warrants additional discussion. The explorer receives a reward if the discriminator reaches a confidence level that exceeds a preset threshold and thus terminates the exploration. However, the reward for the explorer is not conditioned on the correctness of the prediction. This is in keeping with our tenet of separating the exploration from decision making: the explorer is not aware of prediction correctness and it is rewarded as long as the discriminator is confident enough to make a prediction.

\subsection{Co-training}

\begin{figure}
\centering
\includegraphics[width=\linewidth]{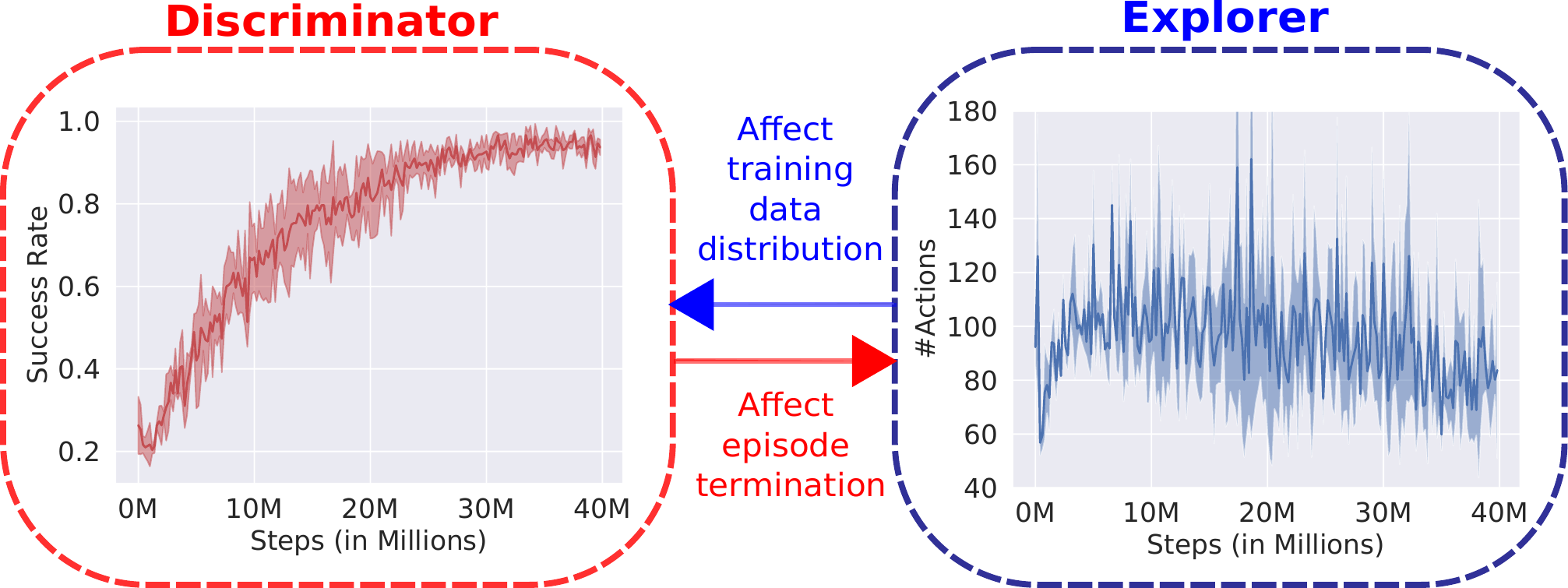}
\caption{Training plots of the discriminator and explorer, and illustration of how they affect each other in the co-training process. The left and right plots show the success rate and the number of actions over the last 100 episodes. Results are averaged over three random seeds and one standard deviation is shaded.}
\label{fig:relationship}
\vspace{-0.15in}
\end{figure}

\remindtext{1-6}{While our architecture is constructed around separate discriminator and explorer modules, we find that the interplay and inter-dependencies between the two components make independent training infeasible and suggest a co-training framework. On one hand, training the discriminator requires a labeled dataset with partial observations of object geometry, but the distribution of partial observability highly depends on the exploration policy. On the other hand, training the explorer needs termination signals provided by the discriminator. This termination signal can highly affect the explorer's learning efficiency.} \addedtext{1-6b}{Co-training is also important because any pre-trained discriminator will not generalize well as the explorer evolves and implicitly changes the distribution of the data presented to the discriminator. To handle this shift, the discriminator needs to co-evolve with the explorer.}



Our co-training process is shown in Alg.~\ref{algo:ours}. Initially, both discriminator and explorer are initialized randomly. We collect an initial data buffer of labeled samples for the discriminator with a randomly initialized explorer. In the co-training loop, we first train the discriminator using the data buffer. Then we fix the discriminator, train the explorer, and, at the same time, push the partially observed occupancy grids collected by the explorer along with their ground truth identities into the data buffer. The updated data buffer is used for discriminator training in the next iteration. 

\begin{algorithm}
\SetAlgoLined
 Initialize discriminator randomly\;
 Initialize explorer randomly\;
 Collect an initial data buffer $\mathcal{D}$ using the explorer\;
 \While{steps $<$ maximum step}{
  Train the discriminator for $N_d$ epochs\;
  Fix the discriminator, train the explorer for $N_e$ steps, and push all occupancy grids (with object identity labels) collected by the explorer into data buffer $\mathcal{D}$\;
 }
 \caption{Co-training Discriminator and Explorer}
 \label{algo:ours}
\end{algorithm}

In this process, the discriminator affects episode termination and the explorer affects partial observability of the labeled training data (Fig.~\ref{fig:relationship}). The explorer is rewarded when the discriminator becomes certain and terminates the episode; thus, it learns to make the discriminator confident as quickly as possible. Batch training of the discriminator with samples collected by the explorer also facilitates data reuse and efficiency. Every time one component gets improved, the other component adapts to the distributional shift. Because updates happen with each iteration, this shift is manageable. As a result, the discriminator and the explorer co-evolve, gradually pushing the other to improve and eventually converge. 

\section{Experiments}
\begin{table*}
    \caption{Comparative performance of various methods in simulation under 0.1\% and 0.5\% sensor failure rate. For each method, we present the number of actions taken (\#Actions) and the number of pixels explored  (\#Explored Pixels) before making a prediction, as well as the success rate in identifying the correct object (Success Rate). Mean and standard deviation over 1,000 trials are shown. A detailed description of each method can be found in Sec.~\ref{sec:methods}.}
    \label{tab:results}
    \centering
    \begin{tabular}{c|ccc|ccc}
    \toprule
    \multirow{2}{*}{\textbf{Methods}} & \multicolumn{3}{c|}{\textit{0.1\% Sensor Failure}} & \multicolumn{3}{c}{\textit{0.5\% Sensor Failure}} \\
    & \#\textbf{Actions} & \textbf{\#Explored Pixels} & \textbf{Success Rate} & \#\textbf{Actions} & \textbf{\#Explored Pixels} & \textbf{Success Rate} \\
    \midrule
    Random-walk & 1427 $\pm$ 654.8 & 354.8 $\pm$ 148.9 & 0.31 & 1350 $\pm$ 667.4 & 338.3 $\pm$ 148.5 & 0.27 \\
    Not-go-back & 684.5 $\pm$ 565.9 & 466.6 $\pm$ 320.4 & 0.49 & 621.4 $\pm$ 524.7 & 427.9 $\pm$ 293.8 & 0.43 \\
    Info-gain & 435.1 $\pm$ 397.5 & 341.7 $\pm$ 250.3 & 0.45 & 365.1 $\pm$ 360.6 & 291.2 $\pm$ 232.2 & 0.42 \\   
    Edge-follower & 60.05 $\pm$ 218.6 & 33.01 $\pm$ 15.95 & 0.91 & 95.24 $\pm$ 282.5 & 32.48 $\pm$ 32.81 & 0.75 \\
    Edge-ICP & 136.1 $\pm$ 339.1 & 72.29 $\pm$ 16.78 & 0.94 & 400.6 $\pm$ 719.4 & 75.63 $\pm$ 41.35 & 0.81 \\
    PPO-ICP& 921.2 $\pm$ 679.1 & 286.2 $\pm$ 189.6 & 0.35 & 860.4 $\pm$ 698.3 & 231.7 $\pm$ 172.4 & 0.31 \\
    All-in-one & 28.63 $\pm$ 207.8 & 3.827 $\pm$ 6.735 & 0.23 & 66.05 $\pm$ 328.0 & 6.229 $\pm$ 15.15 & 0.22 \\
    TANDEM (ours) & 54.97 $\pm$ 106.5 & 44.74 $\pm$ 37.32 & 0.96 & 64.76 $\pm$ 109.3 & 49.71 $\pm$ 36.27 & 0.95 \\  
    \bottomrule
    \end{tabular}
    \vspace{-0.1in}
\end{table*}

In this section, we describe our experimental setup, in both simulation and the real world\footnote{For real-world video demonstrations or more information, please visit our project website at \url{https://jxu.ai/tandem}.}. Our method is trained entirely in simulation; it can then be tested either in simulation or on a real robot. We present an extensive set of comparisons against a number of baselines in simulation, then validate the performance of our method on real hardware. 

\subsection{Setup}

Our experiments assume a tactile finger that moves on a 30cm by 30cm plane and is always perpendicular to the plane (Fig.~\ref{fig:teaser}). The target object is placed roughly at the center of the workspace in any random orientation. The object is fixed and does not move after interaction with the finger. At each time step $t$, the robot can execute an action $a_t \in \mathcal{A} = \{\text{up, right, down, left}\}$ which corresponds to a 5mm translation in the 4 directions on the plane. \remindtext{1-1b}{After each action, the robot receives a binary collision signal $s_t \in \{0, 1\}$, where 0 indicates collision and 1 indicates collision-free. As described above, this information is encoded in an occupancy grid with a 5mm cell size.}

In real-world experiments, we use the DISCO finger~\cite{piacenza2020sensorized} as our tactile sensor (Fig.~\ref{fig:teaser}), but discard additional tactile information (such as contact force magnitude) and only rely on touch/no-touch data. We mount the finger on a UR5 robot arm. For simulation, we use the PyBullet engine and assume a floating finger with similar tactile capabilities. 

Sensor noise is an important consideration since most real-world tactile sensors exhibit some level of noise in their readings, and ours is no exception. It is important for any tactile-based methods to be able to handle erroneous readings without compromising efficiency or accuracy. In particular, we found through empirical observations of our sensors that the chance of an incorrect touch signal being reported is around 0.3\% - 0.5\%. We thus compared all the methods presented below for relevant levels of tactile sensor noise. For learning-based methods, we also have the option of simulating noise during the training process in order to increase robustness; in our case, we simulate a 0.5\% sensor failure rate in the co-training process for our method. 

We generate 10 polygons with random shapes as our test objects, as shown in Fig.~\ref{fig:teaser}. These polygons are generated by walking around the circle, taking a random angular step each time, and at each step putting a point at a random radius. The maximum number of edges is 8 and the maximum radius for each sampled point is 10cm. We 3D-print these polygons for real-world experiments or use their triangular meshes for the simulated versions. For simulation, we decompose each polygon into a set of convex parts for collision checking.

Each episode is terminated when the confidence of the discriminator is greater than the preset threshold of 0.98 or the number of actions has exceeded 2,000. At termination, the prediction of the discriminator is compared to the ground truth identification of that object to check success.

\subsection{Training}

We train our proposed method entirely in simulation. In each co-training iteration, the discriminator is trained for $N_d=15$ epochs on the data buffer of size $|\mathcal{D}|=1e^{6}$, and the explorer is trained for $N_e=2e^5$ steps.  A 0.5\% sensor failure noise is applied during training.

Fig.~\ref{fig:relationship} shows the training plots during our co-training process. Our method's ability to correctly recognize the object (success rate) improves consistently during the process; however, the number of actions taken for the explorer to make the discriminator confident starts at a low level, first increases, and then drops after peaking at around 5M steps. Our discriminator is initialized randomly and when the training starts, it is making bold decisions to terminate the exploration quickly. This is why the number of actions starts low and the success rate is also bad in the beginning. However, as more and more labeled counter-examples of such wrong termination are gathered by the explorer and added to the data buffer of the discriminator, the discriminator starts to become cautious, and thus the number of actions to make it confident grows. At around 5M steps,  a decent enough discriminator is obtained for the explorer and discriminator to start co-evolving until convergence.

\subsection{Baselines}
\label{sec:methods}

In order to evaluate the effectiveness of our learned exploration policy on the tactile object recognition task, we choose to compare our approach to learned all-in-one (without separating exploration and discrimination) and non-learned (heuristic-based) baselines. The metrics that we are most interested in are the number of actions and the success rate in accurately identifying the objects. The methods we evaluate are as follows: 

\subsubsection{Random-walk} This method generates a random move at each step. A discriminator is trained with this exploration policy for object identification and terminating exploration. We apply a 0.5\% sensor failure rate during training. 

\subsubsection{Not-go-back} Similar to \textit{Random-walk}, except that the random move generated at each time step is always to an unexplored neighboring pixel.

\subsubsection{Info-gain} This method uses the info-gain heuristics: it also picks an action that leads to an unexplored pixel, but, unlike \textit{Not-go-back} which picks it randomly, it picks the action that provides the most salient information. At time step $t$, let $\mathbf{p}$ denote the probability distribution over 10 objects predicted by the discriminator on the current grid. Let $\mathbf{p_w}$ and $\mathbf{p_b}$ denote the new probability distributions if the newly explored pixel turns out to be white and black respectively, after applying a particular action. Then the action $a_t$ is chosen by:
\begin{equation*}
    a_t = \argmax_{a \in \mathcal{A}} \: \bigg\{ \mathcal{H} (\mathbf{p}) - \left(\frac{1}{2} \mathcal{H} (\mathbf{p_w}) + \frac{1}{2} \mathcal{H} (\mathbf{p_b})\right) \bigg\}
\end{equation*}
where $\mathcal{H}$ denotes the entropy of a probability distribution. It uses entropy as a measure of uncertainty and picks an action that provides the most information gain (reduces the most uncertainty). A discriminator is trained and we apply a 0.5\% sensor failure rate during training. 

\begin{figure}[t!]
\centering
\includegraphics[width=\linewidth]{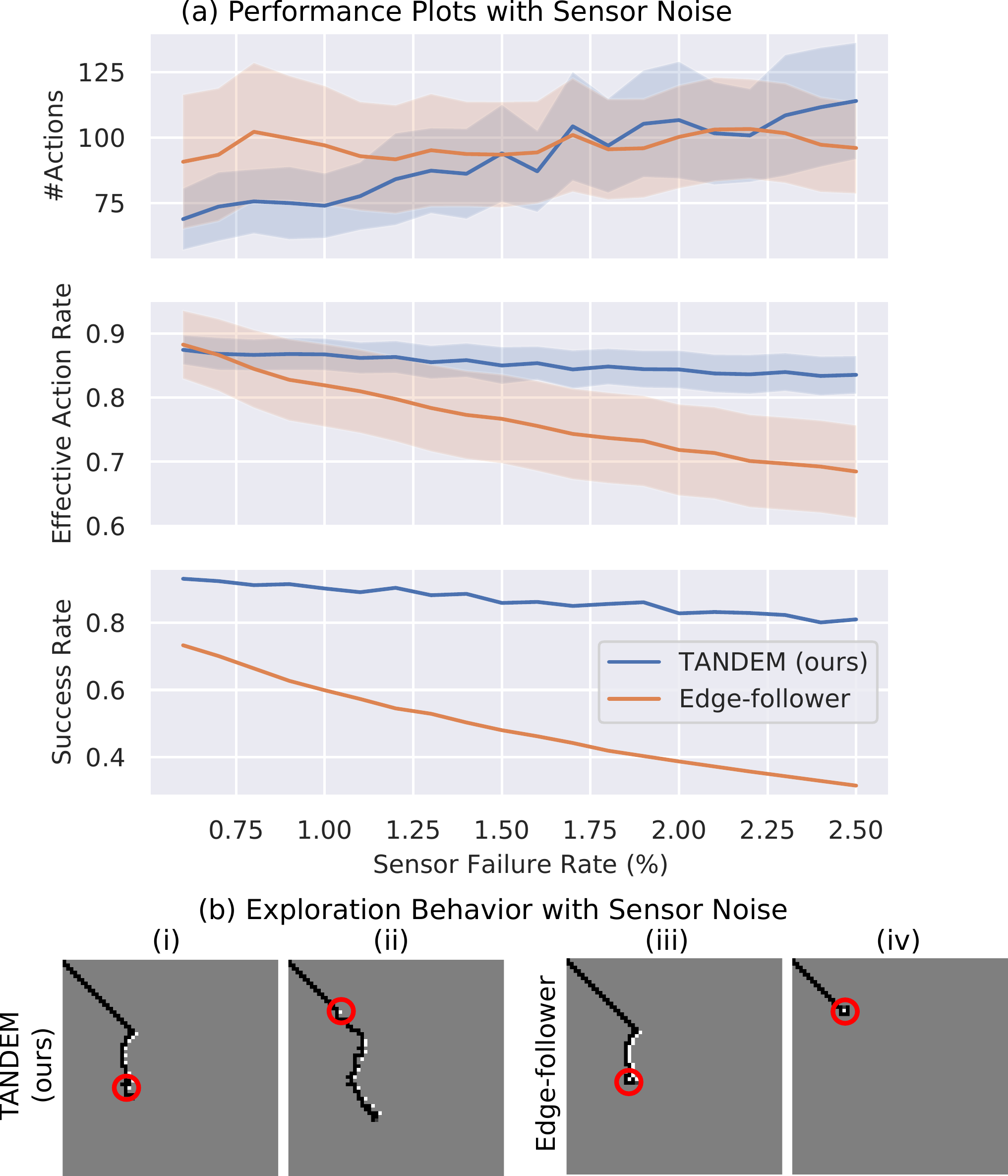}
\caption{(a) Performance of \textit{TANDEM} and \textit{Edge-follower} as the sensor failure rate increases from 0.6\% to 2.5\%. For \#Actions, $\pm$ 0.1 standard deviation is shaded. For Effective Action Rate, $\pm$ 0.2 standard deviation is shaded. With higher sensor noise, both methods need more actions. However, \textit{TANDEM} retains a high success rate and action efficiency while those of \textit{Edge-follower} deteriorate continuously. (b) Exploration behavior of \textit{TANDEM} and \textit{Edge-follower} when sensor failure happens. The location of the sensor failure is circled in red (in the simulation we can ensure it occurs at the same location for both methods). (i)(iii) show a sensor failure after contacting object 1, and (ii)(iv) show a sensor failure before contacting object 5. For these two examples, \textit{Edge-follower} makes the wrong prediction with 39 and 6 actions while \textit{TANDEM} correctly identifies the objects with 38 and 79 actions respectively.}
\label{fig:noise_demo}
\vspace{-0.15in}
\end{figure}

\subsubsection{Edge-follower} This method uses the popular contour-following heuristic as the exploration policy. A discriminator is trained in this method but we do not apply sensor noise during training. We notice that when applying sensor noise during training, the performance of the \textit{Edge-follower} drops significantly. This is because \textit{Edge-follower} can sometimes get trapped at locations where a collision-free pixel is identified as collision and starts circling that pixel. In such a case, unlike other methods such as \textit{Random} and \textit{Not-go-back}, the \textit{Edge-follower} can not keep exploring with random actions. Thus, the discriminator trained in \textit{Edge-follower} becomes unnecessarily cautious but its exploration policy is not able to increase its confidence. 

\subsubsection{Edge-ICP} This method uses the same exploration policy as \textit{Edge-follower}. However, instead of training a learning-based discriminator, it uses the Iterative Closest Point (ICP) algorithm. The occupancy grid is converted to a point cloud using the center location of each pixel. The discriminator runs ICP to match the point cloud to each object using 36 different initial orientations evenly spaced between [0\degree, 360\degree]. For each object, the minimum error among all orientations represents the matching quality. If the error is smaller than 0.0025cm then the object is marked as a match. The output probability distribution assigns equal probabilities to the matched objects and zeroes to not-matched ones. There is no training required for this method.

\subsubsection{PPO-ICP} This method trains a PPO explorer using the ICP discriminator as in \textit{Edge-ICP}. A 0.5\% sensor failure rate is applied during training. 

\subsubsection{All-in-one} This method does not separate explorer and discriminator. It has the same structure as the PPO explorer proposed in our approach except that the action space has been expanded to 14 actions. The first 4 actions correspond to a move and the remaining 10 actions correspond to a prediction. If a prediction is made, the episode is terminated. A reward of 1 is given only when the episode is terminated and the prediction is correct. A 0.5\% sensor failure rate is applied during training.

\subsubsection{TANDEM} This is our proposed method. 

\begin{figure*}[h!]
\centering
\includegraphics[width=\textwidth]{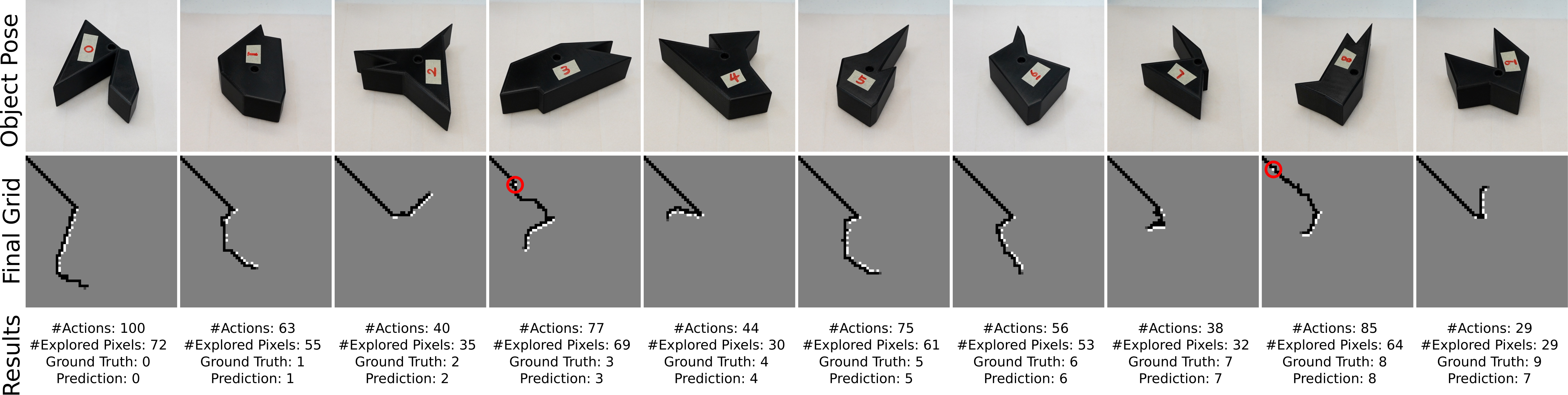}
\caption{10 examples of our method on real robot experiments. The top row shows the object poses, the medium row shows the occupancy grids at termination, and the last row shows the results for each trial. The first 9 examples are successful and the last one is a failure case. While sensor noise can happen anywhere in a trial, it is easier to identify when it occurs before the contact. We highlight in red circles such sensor noise for objects 3 and 8. Our method is able to bypass the noisy pixel,  continue exploring and make the correct prediction.}
\label{fig:real}
\vspace{-0.15in}
\end{figure*}

\begin{figure}[h!]
    \centering
    \includegraphics[width=\linewidth]{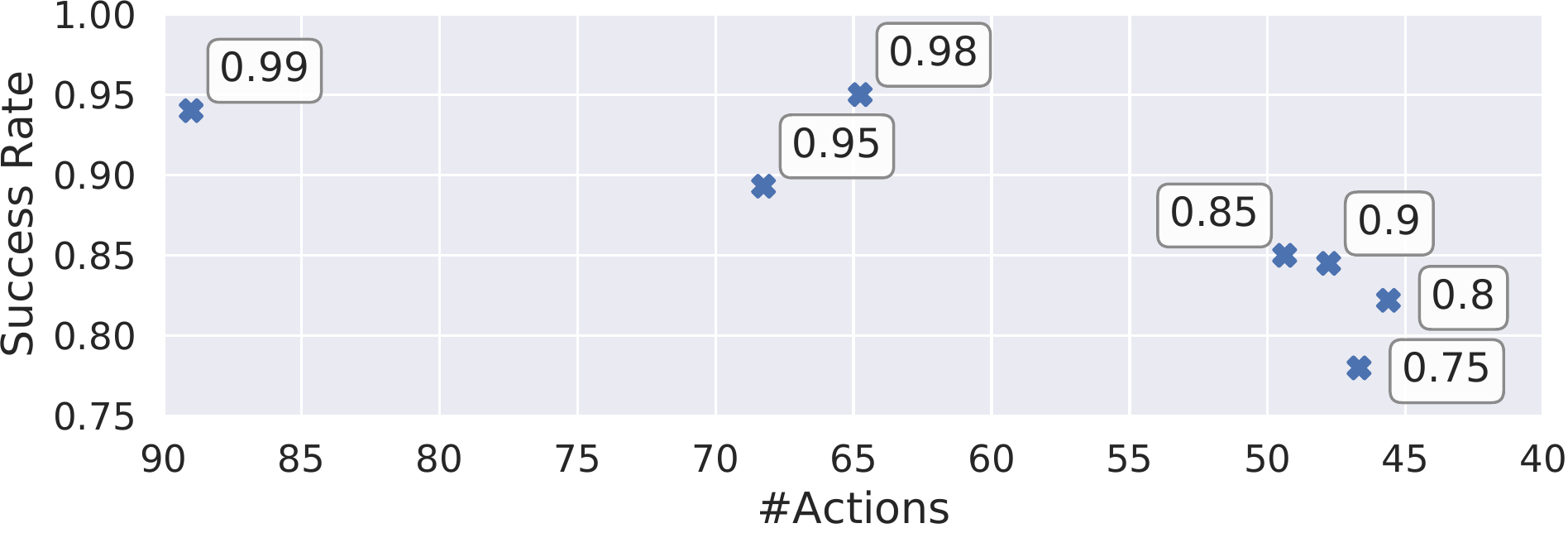}
    \caption{Success Rate and \#Actions when using different threshold values (annotated in text box) in co-training. Each number is computed with 1,000 trials. Note that a threshold of 1 is too strict and the models converge at around 1,500 actions and a 0.4 success rate.}
    \label{fig:thresholds}
    \vspace{-0.15in}
\end{figure}  

\subsection{Comparative Performance Analysis} We compare all methods described above over a large set of simulated experiments, as shown in Table~\ref{tab:results}. 

For both sensor noise levels we consider, \textit{TANDEM} outperforms the baselines in terms of both success rate and the number of actions required. Only \textit{All-in-one} uses fewer actions at 0.1\% sensor noise but at the price of an extremely low success rate. 

We attribute this gap in performance to multiple factors. For example, while \textit{Random-walk} or \textit{Not-go-back} are clearly inefficient exploration strategies, \textit{Info-gain} is a popular heuristic-based method and has been shown to be efficient in other contexts by previous works. However, we found it to not work well in conjunction with a CNN discriminator. Compared to other methods, the \textit{Info-gain} explorer is more dependent on the discriminator because the discriminator affects not only the termination of each episode but also the action selection at each time step. For the \textit{Info-gain} explorer to be effective, it likely requires a discriminator with high accuracy to begin with, which our method does not. The \textit{All-in-one} method, which is not equipped with a dedicated discriminator, cannot train decision-making directly using the labeled samples collected by the explorer, leading to inefficient training and much worse performance if given the same amount of training time as \textit{TANDEM}.

Edge-following, unsurprisingly, is an efficient exploration heuristics for our task, given its 2D nature. \textit{Edge-follower} and \textit{Edge-ICP} have the best performance among all baselines. However, they are shown to be very sensitive to sensor noise, in terms of both accuracy and efficiency. To further investigate this aspect, we compared \textit{TANDEM} and \textit{Edge-follower} for sensor failure chance further increased up to 2.5\%. As shown in Fig.~\ref{fig:noise_demo}, despite being trained with a fixed 0.5\% sensor noise, \textit{TANDEM} maintains a high success rate even in the presence of more noise. We also report the Effective Action Rate (EAR) in this experiment, where EAR is computed as \#Explored Pixels / \#Actions per episode, a metric reflecting the effectiveness of the move in exploring new locations. We can see that the actions generated by our method maintain high exploration efficiency as shown by the EAR plot. In comparison, both EAR and success rate drop as the sensor failure rate increases for \textit{Edge-follower}. Both methods need longer episode lengths to handle larger sensor noise. Two examples of exploration behavior under noise are shown in Fig.~\ref{fig:noise_demo}. \textit{Edge-follower} makes the wrong prediction for both examples while \textit{TANDEM} successfully handles both. This is due to \textit{Edge-follower}'s discrimination policy overfitting to the edge-following behavior and not being able to explore further after being trapped at an incorrect collision signal. 

Unlike \textit{Edge-ICP}, \textit{PPO-ICP} struggles to achieve similar performance. ICP needs a sufficient number of points to achieve decent recognition accuracy and terminate the exploration because it is not able to utilize non-collision pixels. While the edge-following policy is good at collecting points through constantly touching the object, the PPO explorer struggles at learning similar behavior because of the extremely sparse termination reward provided by ICP.

\subsection{Confidence Threshold}
\addedtext{1-5}{The confidence threshold used by the discriminator to determine termination has a large effect on the performance of the co-training framework. Our threshold value of 0.98 is chosen empirically. Fig.~\ref{fig:thresholds} shows the number of actions and success rate with different thresholds used in co-training. Smaller confidence thresholds make the discriminator terminate the exploration earlier. Thus, when the co-training converges, fewer actions are needed but at the same time, the success rate of correctly identifying the objects is worse. We choose 0.98 because it achieves a good trade-off between the success rate ($\geq 0.95$) and the number of actions ($\leq 65$).}

\subsection{Real-World Performance}

We validate the performance of \textit{TANDEM} on a real robot. We run 3 trials for each of 10 objects with random orientations (30 trials total), with results shown in Table~\ref{tab:real}.

Our method still achieves a high identification accuracy, even if slightly lower when compared to simulation results at a 0.5\% sensor failure rate. Exploration efficiency, as illustrated by the number of actions, is at similar levels. We attribute the sim-to-real gap to imperfections in our noise models, shape printing, and robot control.

Fig.~\ref{fig:real} shows ten examples of \textit{TANDEM} in operation, one for each object in a random orientation, also showing the occupancy grid at the moment that a decision is made. This decision is correct 90\% of the time despite the limited nature of the information collected by that point. We also note that our method is robust enough to handle sensor noise, even before making first contact (objects 3 and 8). We also show a failure case where our method incorrectly recognizes object 9 as object 7: both these polygons have a large opening triangle, which makes them hard to distinguish when this area is under contact. Our learned exploration policy is often similar to edge-following, but has the added ability to handle sensor noise, and also learns to take shortcuts when appropriate and take advantage of non-collision pixels for discrimination: the discriminator terminates the episode at a non-collision location for object 0.

\begin{table}
    \setlength\tabcolsep{4.5pt}
    \caption{Real robot experiment results (mean and standard deviation over 30 trials).}
    \label{tab:real}
    \centering
    \begin{tabular}{c|ccc}
    \toprule
    \textbf{Method} & \#\textbf{Actions} & \textbf{\#Explored Pixels} &  \textbf{Success Rate} \\
    \midrule
    TANDEM & 67.33 $\pm$ 23.47 & 53.95 $\pm$ 18.16 & 0.90 (27/30) \\  
    \bottomrule
    \end{tabular}
    \vspace{-0.15in}
\end{table}
\section{Conclusion}

We present TANDEM, a new architecture to learn active and efficient exploration policy with task-related decision making. Our approach consists of distinct modules for exploration, discrimination, and world encoding. Even though our approach separates exploration and discrimination, they are co-trained interweavingly. The explorer learns to reveal useful information to the discriminator efficiently and the discriminator adapts to the partial observability of labeled data collected by the explorer. We show that they co-evolve and converge at the end of the training process. We demonstrate our method on tactile object recognition and compare our approach against multiple baselines for exploration (such as edge-following and info-gain) and recognition (such as ICP). Our experiments show that TANDEM recognizes objects with a higher success rate and lower number of movements. Our real-robot experiments demonstrate that our approach, despite being trained purely in simulation, transfers well to the real hardware, and is robust to sensor noise. \addedtext{1-7}{Future directions include generalizing to high-dimensional tactile data and extending our framework to also estimate object orientations and locations along with object identities.}

\bibliographystyle{IEEEtran}
\bibliography{references}

\begin{thebibliography}{10}
\providecommand{\url}[1]{#1}
\csname url@rmstyle\endcsname
\providecommand{\newblock}{\relax}
\providecommand{\bibinfo}[2]{#2}
\providecommand\BIBentrySTDinterwordspacing{\spaceskip=0pt\relax}
\providecommand\BIBentryALTinterwordstretchfactor{4}
\providecommand\BIBentryALTinterwordspacing{\spaceskip=\fontdimen2\font plus
\BIBentryALTinterwordstretchfactor\fontdimen3\font minus
  \fontdimen4\font\relax}
\providecommand\BIBforeignlanguage[2]{{%
\expandafter\ifx\csname l@#1\endcsname\relax
\typeout{** WARNING: IEEEtran.bst: No hyphenation pattern has been}%
\typeout{** loaded for the language `#1'. Using the pattern for}%
\typeout{** the default language instead.}%
\else
\language=\csname l@#1\endcsname
\fi
#2}}

\bibitem{okamura2001feature}
A.~M. Okamura and M.~R. Cutkosky, ``Feature detection for haptic exploration
  with robotic fingers,'' \emph{The Intl. Journal of Robotics Research},
  vol.~20, no.~12, pp. 925--938, 2001.

\bibitem{allen1985object}
P.~K. Allen, ``Object recognition using vision and touch,'' 1985.

\bibitem{meier2011probabilistic}
M.~Meier, M.~Schopfer, R.~Haschke, and H.~Ritter, ``A probabilistic approach to
  tactile shape reconstruction,'' \emph{IEEE Transactions on Robotics},
  vol.~27, no.~3, pp. 630--635, 2011.

\bibitem{allen1988integrating}
P.~K. Allen, ``Integrating vision and touch for object recognition tasks,''
  \emph{The Intl. Journal of Robotics Research}, vol.~7, no.~6, pp. 15--33,
  1988.

\bibitem{bierbaum2009grasp}
A.~Bierbaum, M.~Rambow, T.~Asfour, and R.~Dillmann, ``Grasp affordances from
  multi-fingered tactile exploration using dynamic potential fields,'' in
  \emph{2009 9th IEEE-RAS Intl. Conf. on Humanoid Robots}.\hskip 1em plus 0.5em
  minus 0.4em\relax IEEE, 2009, pp. 168--174.

\bibitem{gaston1984tactile}
P.~C. Gaston and T.~Lozano-Perez, ``Tactile recognition and localization using
  object models: The case of polyhedra on a plane,'' \emph{IEEE transactions on
  pattern analysis and machine intelligence}, no.~3, pp. 257--266, 1984.

\bibitem{skiena1989problems}
S.~Skiena, ``Problems in geometric probing,'' \emph{Algorithmica}, vol.~4,
  no.~4, pp. 599--605, 1989.

\bibitem{piacenza2020sensorized}
P.~Piacenza, K.~Behrman, B.~Schifferer, I.~Kymissis, and M.~Ciocarlie, ``A
  sensorized multicurved robot finger with data-driven touch sensing via
  overlapping light signals,'' \emph{IEEE/ASME Transactions on Mechatronics},
  vol.~25, no.~5, pp. 2416--2427, 2020.

\bibitem{liu2017recent}
H.~Liu, Y.~Wu, F.~Sun, and D.~Guo, ``Recent progress on tactile object
  recognition,'' \emph{Intl. Journal of Advanced Robotic Systems}, vol.~14,
  no.~4, p. 1729881417717056, 2017.

\bibitem{schmitz2014tactile}
A.~Schmitz, Y.~Bansho, K.~Noda, H.~Iwata, T.~Ogata, and S.~Sugano, ``Tactile
  object recognition using deep learning and dropout,'' in \emph{2014 IEEE-RAS
  Intl. Conf. on Humanoid Robots}, pp. 1044--1050.

\bibitem{strub2014using}
C.~Strub, F.~W{\"o}rg{\"o}tter, H.~Ritter, and Y.~Sandamirskaya, ``Using
  haptics to extract object shape from rotational manipulations,'' in
  \emph{2014 Intl. Conf. on Intelligent Robots and Systems}, pp. 2179--2186.

\bibitem{pastor2019using}
F.~Pastor, J.~M. Gandarias, A.~J. Garc{\'\i}a-Cerezo, and J.~M. G{\'o}mez-de
  Gabriel, ``Using 3d convolutional neural networks for tactile object
  recognition with robotic palpation,'' \emph{Sensors}, vol.~19, no.~24, p.
  5356, 2019.

\bibitem{allen1988haptic}
P.~K. Allen and K.~S. Roberts, ``Haptic object recognition using a
  multi-fingered dextrous hand,'' 1988.

\bibitem{watkins2019multi}
D.~Watkins-Valls, J.~Varley, and P.~Allen, ``Multi-modal geometric learning for
  grasping and manipulation,'' in \emph{2019 Intl. Conf. on robotics and
  automation}.\hskip 1em plus 0.5em minus 0.4em\relax IEEE, 2019, pp.
  7339--7345.

\bibitem{martinez2013active}
U.~Martinez-Hernandez, G.~Metta, T.~J. Dodd, T.~J. Prescott, L.~Natale, and
  N.~F. Lepora, ``Active contour following to explore object shape with robot
  touch,'' in \emph{2013 World Haptics Conf. (WHC)}.\hskip 1em plus 0.5em minus
  0.4em\relax IEEE, 2013, pp. 341--346.

\bibitem{yu2015shape}
K.-T. Yu, J.~Leonard, and A.~Rodriguez, ``Shape and pose recovery from planar
  pushing,'' in \emph{2015 IEEE/RSJ Intl. Conf. on Intelligent Robots and
  Systems}.\hskip 1em plus 0.5em minus 0.4em\relax IEEE, 2015, pp. 1208--1215.

\bibitem{suresh2020tactile}
S.~Suresh, M.~Bauza, K.-T. Yu, J.~G. Mangelson, A.~Rodriguez, and M.~Kaess,
  ``Tactile slam: Real-time inference of shape and pose from planar pushing,''
  \emph{arXiv preprint arXiv:2011.07044}, 2020.

\bibitem{pezzementi2011tactile}
Z.~Pezzementi, E.~Plaku, C.~Reyda, and G.~D. Hager, ``Tactile-object
  recognition from appearance information,'' \emph{IEEE Transactions on
  Robotics}, vol.~27, no.~3, pp. 473--487, 2011.

\bibitem{hebert2013next}
P.~Hebert, T.~Howard, N.~Hudson, J.~Ma, and J.~W. Burdick, ``The next best
  touch for model-based localization,'' in \emph{2013 IEEE Intl. Conf. on
  Robotics and Automation}.\hskip 1em plus 0.5em minus 0.4em\relax IEEE, 2013,
  pp. 99--106.

\bibitem{xu2013tactile}
D.~Xu, G.~E. Loeb, and J.~A. Fishel, ``Tactile identification of objects using
  bayesian exploration,'' in \emph{2013 IEEE Intl. Conf. on Robotics and
  Automation}.\hskip 1em plus 0.5em minus 0.4em\relax IEEE, 2013, pp.
  3056--3061.

\bibitem{schneider2009object}
A.~Schneider, J.~Sturm, C.~Stachniss, M.~Reisert, H.~Burkhardt, and W.~Burgard,
  ``Object identification with tactile sensors using bag-of-features,'' in
  \emph{2009 IEEE/RSJ Intl. Conf. on Intelligent Robots and Systems}.\hskip 1em
  plus 0.5em minus 0.4em\relax IEEE, 2009, pp. 243--248.

\bibitem{driess2017active}
D.~Driess, P.~Englert, and M.~Toussaint, ``Active learning with query paths for
  tactile object shape exploration,'' in \emph{2017 IEEE/RSJ Intl. Conf. on
  Intelligent Robots and Systems}.\hskip 1em plus 0.5em minus 0.4em\relax IEEE,
  2017, pp. 65--72.

\bibitem{rajeswar2021touch}
S.~Rajeswar, C.~Ibrahim, N.~Surya, F.~Golemo, D.~Vazquez, A.~Courville, and
  P.~O. Pinheiro, ``Touch-based curiosity for sparse-reward tasks,''
  \emph{arXiv preprint arXiv:2104.00442}, 2021.

\bibitem{zhang2017active}
M.~M. Zhang, N.~Atanasov, and K.~Daniilidis, ``Active end-effector pose
  selection for tactile object recognition through monte carlo tree search,''
  in \emph{2017 IEEE/RSJ Intl. Conf. on Intelligent Robots and Systems}.\hskip
  1em plus 0.5em minus 0.4em\relax IEEE, 2017, pp. 3258--3265.

\bibitem{fishel2012bayesian}
J.~A. Fishel and G.~E. Loeb, ``Bayesian exploration for intelligent
  identification of textures,'' \emph{Frontiers in neurorobotics}, vol.~6,
  p.~4, 2012.

\bibitem{lepora2013active}
N.~F. Lepora, U.~Martinez-Hernandez, and T.~J. Prescott, ``Active touch for
  robust perception under position uncertainty,'' in \emph{2013 IEEE Intl.
  Conf. on Robotics and Automation}.\hskip 1em plus 0.5em minus 0.4em\relax
  IEEE, 2013, pp. 3020--3025.

\bibitem{martinez2017active}
U.~Martinez-Hernandez, T.~J. Dodd, M.~H. Evans, T.~J. Prescott, and N.~F.
  Lepora, ``Active sensorimotor control for tactile exploration,''
  \emph{Robotics and Autonomous Systems}, vol.~87, pp. 15--27, 2017.

\bibitem{kaboli2017tactile}
M.~Kaboli, D.~Feng, K.~Yao, P.~Lanillos, and G.~Cheng, ``A tactile-based
  framework for active object learning and discrimination using multimodal
  robotic skin,'' \emph{IEEE Robotics and Automation Letters}, vol.~2, no.~4,
  pp. 2143--2150, 2017.

\bibitem{kaboli2019tactile}
M.~Kaboli, K.~Yao, D.~Feng, and G.~Cheng, ``Tactile-based active object
  discrimination and target object search in an unknown workspace,''
  \emph{Autonomous Robots}, vol.~43, no.~1, pp. 123--152, 2019.

\bibitem{schulman2017proximal}
J.~Schulman, F.~Wolski, P.~Dhariwal, A.~Radford, and O.~Klimov, ``Proximal
  policy optimization algorithms,'' \emph{arXiv preprint arXiv:1707.06347},
  2017.

\end{thebibliography}

\end{document}